\pdfoutput=1
\documentclass[11pt]{article}

\usepackage{acl}

\usepackage{times,latexsym}
\usepackage{url}
\usepackage[T1]{fontenc}
\usepackage{graphicx}
\usepackage{times}
\usepackage{latexsym}
\usepackage{amsmath}
\usepackage{longtable}
\usepackage{booktabs}
\usepackage{multirow}
\usepackage[utf8]{inputenc}
\usepackage[T1]{fontenc}
\usepackage[scaled=0.8]{beramono}
\usepackage{microtype}
\usepackage[normalem]{ulem}
\usepackage{enumitem}
\usepackage{inconsolata}

\usepackage{tikz}
\usetikzlibrary{calc}
\usetikzlibrary{decorations.pathmorphing}
\usetikzlibrary{shapes,arrows,chains}
\usetikzlibrary{shapes.geometric}
\usetikzlibrary{patterns}
\usetikzlibrary{positioning}
\tikzset{>=latex}
\usepackage{pgfplots}
\usepackage{relsize}
\title{GMNLP at SemEval-2023 Task 12: \\ Sentiment Analysis with Phylogeny-Based Adapters}

\author{Md Mahfuz Ibn Alam$^*$  ~\  Ruoyu Xie$^*$  ~\  Fahim Faisal$^*$  ~\  Antonios Anastasopoulos \\
        Department of Computer Science, George Mason University \\ 
\texttt{\{malam21, rxie, ffaisal, antonis\}@gmu.edu} }

\begin{document}

\maketitle
\begin{abstract}
This report describes GMU’s sentiment analysis system for the SemEval-2023 shared task AfriSenti-SemEval. We participated in all three sub-tasks: Monolingual, Multilingual, and Zero-Shot. Our approach uses models initialized with AfroXLMR-large, a pre-trained multilingual language model trained on African languages and fine-tuned correspondingly. We also introduce augmented training data along with original training data. Alongside fine-tuning, we perform phylogeny-based adapter-tuning to create several models and ensemble the best models for the final submission. Our system achieves the best F$_1$-score on track~5: Amharic, with~6.2 points higher F$_1$-score than the second-best performing system on this track. Overall, our system ranks 5\textsuperscript{th} among the~10 systems participating in all~15 tracks.

\def
\thefootnote{*}
\footnotetext{Joint contributions: MA performed model training, RX worked on data processing and paper writing, FF constructed the model.}

\end{abstract}

\section{Introduction}
With the increasing use of the internet and social media, the digital availability of various languages is rapidly expanding. This expansion opens avenues for Natural Language Processing (NLP) applications such as Sentiment Analysis (SA) and Machine Translation (MT). Nevertheless, despite African languages comprising 30\% of around 6,000 living languages \cite{skutnabb2003sharing}, most of them are not supported by modern language technologies, leading to an ever-widening gap in language technology access~\cite{joshi-etal-2020-state,blasi-etal-2022-systematic}.

Recently, SA has gained increasing attention, with its applications in various domains, such as public health, literature, and social sciences \cite{mohammad2022ethics}. Despite the growth in this area, most previous works do not include African languages. This shared task focuses on SA on a Twitter dataset in 14 African languages, including Hausa (ha), Yoruba (yo), Igbo (ig), Nigerian Pidgin (pcm), Amharic (am), Tigrinya (tg), Oromo (or), Swahili (sw), Xitsonga (ts), Algerian Arabic (dz), Kinyarwanda (kr), Twi (twi), Mozambican Portuguese (pt), and Moroccan Darija (ma).

This paper presents a novel SA system that effectively addresses the challenge of low-resource and multilingual sentiment classification for multiple African languages. We leverage multilingual language models and propose data augmentation methods to increase the training data size. In addition, we perform phylogeny-based adapter-tuning to create several models. These models are then ensembled to create the final model.

\begin{table}[t]\centering
\resizebox{0.4\textwidth}{!}{%
\begin{tabular}{c|c|c}
\toprule
\textbf{Ranking} & \textbf{Team Name} & \textbf{Avg. F$_1$}\\
\midrule
1 & BCAI-AIR3 & 69.77\\
2 & UM6P & 68.08\\
3 & mitchelldehaven  & 67.85\\
4 & tmn & 65.84\\ 
\textbf{5} & \textbf{GMNLP}  & \textbf{65.76}\\
6 & UCAS  & 65.62\\
7 & Masakhane-Afrisenti  & 63.58\\
9 & PA14  & 63.09\\
9 & DN  &  64.10\\
10 & FUOYENLP  & 59.46\\ 
\bottomrule
\end{tabular}
}
\caption{We calculate the macro-average F$_1$-score for the 10 systems (out of~47) participating in all 15 tracks in this Shared Task. Overall, our system ranks~5\textsuperscript{th}.}
\label{tab:overall_ranking}
\vspace{-0.5em}
\end{table}

\section{Related Work}
While SA is a popular research area in NLP, its application in African languages is relatively rare. This is mainly due to the lack of resources, making it challenging to obtain the data needed to train and evaluate the models~\cite{mabokela2022sentiment}. One solution is creating resources, such as annotated datasets. However, this requires a significant amount of manual annotation~\cite{shode2022yosm}. Using augmented data to improve performance in low-resource languages is another approach that has been explored for various tasks~\cite[][\textit{inter alia}]{xia-etal-2019-generalized,muhammad-etal-2022-naijasenti,alam-anastasopoulos:2022:WMT, xie-anastasopoulos-23-noisy}, where synthetic data is generated from existing data to increase the size of the training set. 

Leveraging pre-trained multilingual language models is a popular choice for SA in African languages~\cite{dossou-etal-2022-afrolm,martin2021sentiment,alabi-etal-2022-adapting,muhammad-etal-2022-naijasenti,martin2022swahbert}. These language models are trained on a large amount of data from different languages, including African languages, which enables them to capture a wide range of linguistic features and patterns. While these pre-trained models have shown promising results, they are imperfect in handling low-resource languages.

Adapters are designed to adapt a large pre-trained language model to a downstream task, enabling efficient transfer learning~\cite{alabi-etal-2022-adapting,ansell-etal-2021-mad-g}. Phylogeny-based adapters~\cite{faisal-anastasopoulos-2022-phylogeny}, similar to the hierarchical ones of~\citet{chronopoulou-etal-2022-efficient} enable knowledge sharing across related languages from the phylogeny-informed tree hierarchy. Our work builds on this approach to address the challenge of SA in low-resource scenarios, demonstrating that it can effectively adapt a pre-trained multilingual language model to the target language with limited training data.

\section{Task Description}
The AfriSenti-SemEval Shared Task 12~\cite{muhammadSemEval2023} consists of three sub-tasks: Monolingual (Task A), Multilingual (Task B), and Zero-Shot (Task C). The primary objective of this shared task is to determine the sentiment of a tweet in a target language, which could be positive, negative, or neutral. A stronger emotion should be chosen when a tweet exhibits positive and negative sentiments. 

\begin{table}[t]
\centering
\resizebox{0.45\textwidth}{!}{%
\begin{tabular}{c|c|c|c}
\toprule
\textbf{Family} & \textbf{Genus} & \textbf{Lang} & \textbf{size} \\
\midrule
\multirow{4}{*}{Afroasiatic} 
& Ethiopic & am  & 5,985\\
& Chadic & ha & 14,173\\
& Arabic & dz & 1,652\\
& Arabic  & ma & 5,584\\

\midrule
\multirow{5}{*}{Niger–Congo} 
& Volta–Congo & ig  & 10,193\\
& Volta–Congo & yo  & 8,523\\
& Bantu & kr &  3,303\\
& Bantu & sw  & 1,811\\
& Bantu & ts &  805\\
& Central Tano & twi &  3,482\\

\midrule
\multirow{1}{*}{Creole} 
& Creole Portuguese & pcm  & 5,122\\

\midrule
\multirow{1}{*}{Indo-European} 
& Romance & pt &  3,064\\

\bottomrule
\end{tabular}
}
\caption{12 languages in Task A, along with their Language Families, Genera, and training data size (sentences). }
\label{table:lang_info}
\vspace{-1em}

\end{table}

\paragraph{Task A: Monolingual}
Task A aims to determine the sentiment of tweets for each language in a monolingual setting. Table \ref{table:lang_info} shows the detail of each language. Every language is a track for this sub-task, creating Tracks~1 through~12.\footnote{Track~13 through Track~15 were not included in the final competition as the respective datasets for these tracks were not released.}

\paragraph{Task B: Multilingual}
Task B aims to determine the sentiment for tweets using a combination of all training data from the~12 languages in Task A. This sub-task includes only one track, Track~16.

\paragraph{Task C: Zero-Shot}
Task C aims to identify the sentiment of tweets from either Tigrinya or Oromo under a zero-shot setting, i.e., no training data are available for these languages). This sub-task is divided into Track 17 (Tigrinya) and 18 (Oromo).

\section{System Overview}
\subsection{Data Source}
The dataset used in the system is mainly sourced from AfriSenti~\cite{muhammad2023afrisenti}, which is already split into train, dev, and test. In addition, we use PanLex \cite{kamholz-etal-2014-panlex} and Stanford Sentiment Tree Bank (SST) \cite{socher-etal-2013-recursive} for data augmentation. 

\paragraph{Panlex} The goal of Panlex is to facilitate the translation of lexemes between all human languages. A broader lexical and linguistic coverage can be achieved when lexemic translations are used instead of grammatical or corpus data. A total of 20 million lexemes have been recorded in PanLex in 9,000 languages, and 1.1 billion pairwise translations have been recorded.

\paragraph{Stanford Sentiment Tree Bank (SST)} The Stanford Sentiment Tree Bank has a sentiment score from 0 to 1 for each sentence. We labeled the sentence as negative if the score was less than or equal to 0.35. We labeled it neutral if the score was between 0.35 and 0.65, and the rest of the sentences were positive.

\subsection{Data Pre-Processing}
The AfriSenti dataset underwent some pre-processing to remove noise, including \textit{@user} and \textit{RT} handlers, URLs, extra while spaces, multiple consecutive punctuations, and characters.\footnote{For example, "hellooooo" to "helloo".} Emojis were intentionally retained since they are an important part of sentiment analysis, as they often convey emotional content. 

\subsection{Data Augmentation}
To improve the robustness of our language systems to variations in language, we utilize data augmentation techniques to increase the amount of available training data. Specifically, we create three datasets using a dictionary-based augmentation approach, an MT approach, and a combination of the first two approaches. This is just a concatenation of the datasets obtained by the two approaches above.

\paragraph{Dictionary Based} To create more data and handle code-mixed sentences, we employ a naive word-to-word translation augmentation method: 
\begin{itemize}
    \item First, we obtain a Panlex bilingual dictionary from English to a corresponding language.
    \item Second, we have obtained the (English) sentences from the English Stanford Sentiment Tree Bank.
    \item Third, we replace any word from the sentence of the Tree Bank that matches an entry from the dictionary with its translation in the corresponding language.
\end{itemize}

The intuition behind this is that not all English sentence words will be replaced, so it will imitate code-mixing. Also, we anticipate that the translated sentences will largely be ungrammatical, as they are just word-to-word translations with no morphological information, which may lead to word order and morphosyntax errors.
%We hope this "noise" will make our system robust to such noise and code-switching.

\paragraph{Machine Translation Based} We introduce an augmentation technique based on MT. We leveraged the best-performing MT model of~\citet{alam-anastasopoulos:2022:WMT}, which handles almost all of the task's languages, to translate sentences from English Stanford Sentiment Treebank to the corresponding language.

\subsection{Model Overview}
Our system uses a transformer-based multilingual model, AfroXLMR-large \cite{alabi-etal-2022-adapting}. AfroXLMR-large is developed by performing language adaptation of the XLM-R-large model \cite{DBLP:journals/corr/abs-1911-02116} on 17 distinct African languages, including Afrikaans, Amharic, Hausa, Igbo, Malagasy, Chichewa, Oromo, Nigerian-Pidgin, Kinyarwanda, Kirundi, Shona, Somali, Sesotho, Swahili, isiXhosa, Yoruba, and isiZulu, which collectively represent the major African language families. AfroXLMR-large also incorporates three high-resource languages: Arabic, French, and English. When we fine-tune the model for our task, we fine-tune only a task-adapter instead of fine-tuning the whole model.

\begin{figure}[t]
\tikzset{seq/.style={draw=none,fill=gray!20}}
\tikzset{seqblue/.style={draw=none,fill=blue!20}}
\tikzset{seqyellow/.style={draw=none,fill=yellow!20}}
\tikzset{seqgreen/.style={draw=none,fill=green!20}}
\tikzset{seqred/.style={draw=none,fill=red!20}}
\tikzset{seqpurple/.style={draw=none,fill=purple!20}}
\tikzset{seqorange/.style={draw=none,fill=orange!20}}
\tikzset{seqlabel/.style={font={\small}}}
\tikzset{layer/.style={->,thick}}
\centering
\begin{tabular}{c}
\begin{tikzpicture}[x=.9cm,y=0.7cm]
\draw[seq] (-4,-0.25) rectangle (1,0.25);
\node[seqlabel] at (-1.5,0) {\small{Layer $i$}};

\draw[layer] (-1.5,0.3) -- (-1.5,0.6);
\draw[seqpurple] (-4,0.65) rectangle (1,1.15);
\node[seqlabel] at (-1.5,0.9) {\small{Niger Adpt}};

%\draw[layer] (2.5,0.3) -- (2.5,0.7);
%\draw[seqred] (1.2,0.75) rectangle (4,1.25);
%\node[seqlabel] at (2.5,1) {Dravidian Adpt};
%\draw[layer] (2.5,1.3) -- (2.5,1.7);
%\node[seqlabel] at (2.5,2) {\ldots};

\draw[layer] (-2.5,1.2) -- (-2.5,1.5);
\draw[seqgreen] (-4,1.55) rectangle (-1.05,2.05);
\node[seqlabel] at (-2.5,1.8) {\small{Volta}};

\draw[layer] (0,1.2) -- (0,1.5);
\draw[seqorange] (-0.95,1.55) rectangle (1,2.05);
\node[seqlabel] at (0,1.8) {\small{Bantu}};
\draw[layer] (0,2.1) -- (0,2.4);
\node[seqlabel] at (0,2.7) {\ldots};

\draw[layer] (-3.25,2.1) -- (-3.25,2.4);
\draw[seqblue] (-4.25,2.45) rectangle (-2.45,2.95);
\node[seqlabel] at (-3.35,2.7) {\small{Igbo}};

\draw[layer] (-1.5,2.1) -- (-1.5,2.4);
\draw[seqyellow] (-2.35,2.45) rectangle (-0.5,2.95);
\node[seqlabel] at (-1.45,2.7) {\small{Yoruba}};

\draw[layer] (-1.5,3.0) -- (-1.5,3.3);
\draw[layer] (-3.25,3.) -- (-3.25,3.3);

\draw[seq] (-4,3.35) rectangle (1,3.85);
\node[seqlabel] at (-1.5,3.6) {\small{Layer $i+1$}};
\end{tikzpicture}
\end{tabular}
    \caption{Incorporating phylogeny into language models with adapters: %Neural adaptation to new languages: 
    We start with an unadapted model where we
    impose a phylogeny-informed tree hierarchy over adapters.}
    \label{fig:phylogeny}
    \vspace{-1em}
\end{figure}
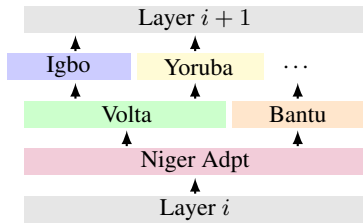

In addition to the base model, we utilize phylogeny-based adapter-tuning~\cite{faisal-anastasopoulos-2022-phylogeny} to generate multiple models for different language families. As described by \citet{houlsby2019parameter}, adapters are small neural components designed to adapt a large pre-trained language model to a downstream task using lightweight layers inserted between each layer of the pre-trained model, enabling efficient adaptation. Here we use phylogeny-based adapters, similar to \citet{faisal-anastasopoulos-2022-phylogeny}, which enables knowledge sharing across similar or related languages from the same language family or genus. Table~\ref{table:lang_info} shows the language family hierarchy for the 12 languages in Task A, and figure~\ref{fig:phylogeny} presents a visualization of the architecture. The language families are chosen based on the phylogenetic relationships among languages, which reflects their evolutionary history and linguistic similarities. For example, the Arabic languages, such as Algerian Arabic and Moroccan Darija, are all derived from the Afroasiatic language family and share many linguistic features. The adapter creates several models we can ensemble to hopefully obtain better results.

\begin{table*}[t]\centering
\resizebox{0.7\textwidth}{!}{%
\begin{tabular}{c|c|c|c|c|c}
\toprule
\textbf{Lang} &\textbf{mbert-cased} &\textbf{mbert-uncased} &\textbf{xlmr} &\textbf{afro-xlmr-large} &\textbf{afri-berta} \\\midrule
am &28.6 &25.8 &58.6 &\textbf{59.5} &54.5 \\
dz &59.6 &58.6 &\textbf{61.7} &41.1 &31.4 \\
ha &71.3 &73.3 &77.5 &\textbf{79.6} &78.4 \\
ig &73 &77.8 &76.1 &\textbf{77.9} &77.1 \\
kr &56.4 &60.7 &61.1 &\textbf{70.8} &60.3 \\
ma &74 &72.4 &77.2 &\textbf{79.5} &60.8 \\
pcm &45.5 &46.2 &48 &\textbf{48.7} &44 \\
pt &60.3 &59.7 &\textbf{74.7} &69.5 &50.3 \\
sw &37 &34.3 &43.6 &\textbf{55.6} &54.1 \\
ts &39.2 &42.5 &34.4 &\textbf{44.3} &39.4 \\
twi &44.3 &45 &\textbf{48.4} &37.8 &46.2 \\
yo &63.3 &63.5 &66 &\textbf{74.5} &70.2 \\\midrule
Average &54.4 &55 &60.6 &\textbf{61.6} &55.6 \\
\bottomrule
\end{tabular}
}
\caption{Weighted F$_1$-score on our unprocessed test set of different pre-trained models.}
\label{tab:mod_dec1}
\end{table*}

\section{Experiments}
\paragraph{Experimental Setup}
We trained our model using the Adam optimizer \cite{kingma2014adam}, with a learning rate of 1$e^{-4}$. The number of epochs was set to 5, and the batch size was 32, with a maximum sequence length of 128. All experiments were conducted on one A100 GPU. We report the weighted F$_1$-score for the model's performance.

\subsection{Model Selection}
We experiment with multiple pre-trained models to decide which pre-trained model works best with the given languages. Here we focus on the 12 languages from Task A. Note that for this analysis, we use the unprocessed data with a slightly different data split, as we did not have access to the gold labels for the dev set. In this experiment, we split the train set into train, dev, and test sets, with proportions of 80\%, 10\%, and 10\%, respectively.

We experiment with the following models
\begin{itemize}[leftmargin=*,noitemsep,nolistsep]
\item \texttt{mBERT} \cite{DBLP:journals/corr/abs-1810-04805}:  a multilingual version of BERT, which is pre-trained on a large amount of multilingual data from Wikipedia.
\item \texttt{XLM-RoBERTa} \cite{DBLP:journals/corr/abs-1911-02116}: a multilingual version of RoBERTa \cite{liu2019roberta}, which is pre-trained on 2.5TB of CommonCrawl data over 100 languages.
\item \texttt{AfriBERTa} \cite{ogueji-etal-2021-small}: a model based on mBERT and (continued) pre-trained on~11 African languages.
\item \texttt{AfroXLMR-large} \cite{alabi-etal-2022-adapting}:  a model that is based on XLM-R-large and pre-trained on~17 African language.
\end{itemize}

We report the performance in Table \ref{tab:mod_dec1} for unprocessed data. Except for Algerian Arabic, Mozambican Portuguese, and Twi, all the other languages have the highest F$_1$-score using the AfroXLMR-large. AfroXLMR-large has the average best score, so we decided to conduct all our further experiments focusing on AfroXLMR-large.

\begin{table*}[t]\centering
\resizebox{0.7\textwidth}{!}{%
\begin{tabular}{c|c|c|c|c}
\toprule
\textbf{Lang} & \textbf{Clean} &\textbf{Clean + Dictionary-based} &\textbf{Clean + MT-based} &\textbf{Clean + Both} \\\midrule
am &62.8 &\textbf{63.6} &61.7 &63.2 \\
dz &58.6 &\textbf{58.6} & -&- \\
ha &\textbf{79.7} &79.2 &78.3 &79.1 \\
ig &73.1 &72.9 &74 &\textbf{74.4} \\
kr &66.5 &\textbf{67.7} &22.4 &66.4 \\
ma &73.3 &\textbf{77.8} & -&- \\
pcm &75.9 &\textbf{76.1} & -&- \\
pt &\textbf{64.4} &- &- &- \\
sw &59 &61.8 &\textbf{62} &60.1 \\
ts &41.7 &39.2 &\textbf{51.3} &44.1 \\
twi &48.5 &\textbf{51.2} & - & -\\
yo &74.7 &74.8 &\textbf{74.9} &74.5 \\
\bottomrule
\end{tabular}
}
\caption{Weighted F$_1$-score on the dev set of our monolingual models on different datasets. The best score per language is \textbf{highlighted}.}
\label{tab:data_dec1}
% \vspace{-1em}
\end{table*}

\subsection{Dataset Selection} 
To decide which dataset works best with the given languages, we create monolingual models with only that language's training data. We test them on the dev set provided by the shared task.~\label{dataset_decision}

\begin{itemize}[leftmargin=*,noitemsep,nolistsep]
\item \texttt{Clean}: All the pre-processed training data from the shared task, without any data augmentation.
\item \texttt{Clean + Dictionary-based}: All the pre-processed training data from the shared task are mixed and shuffled with all the Dictionary-based augmented translation datasets. We did not find any bilingual lexicon on Panlex for Mozambican Portuguese. 
%\an{Did we use Portugal Portuguese then? Or did we not try anything for this language under this setting?}\ma{We did not try anything.}
\item \texttt{Clean + MT-based}: All the pre-processed training data from the shared task are mixed and shuffled with all the MT-based augmented translation datasets. The model we used for translation does not support Twi, Mozambican Portuguese, Nigerian Pidgin, Moroccan Arabic, and Algerian Arabic. Therefore, we do not use the MT data for these languages.
\item \texttt{Clean + Both}: All the pre-processed training data from the shared task are mixed and shuffled with all the MT-based augmented translation and Dict-based augmented translation if both datasets are available for a language.
%\vspace{-1}
\end{itemize}

Table~\ref{tab:data_dec1} shows the result on the dev set when under each of our~4 dataset settings. The first takeaway is that for almost all languages, data augmentation helps. The only exception is Hausa (and Portuguese) which is comparably high-resource. We observe improvements in languages well-supported by the MT system, like Igbo, Swahili, Xitsonga, or Yoruba. Table~\ref{tab:data_dec1} also shows which dataset type works best for a certain language. 

We use that information to compile a combined dataset termed "\texttt{Best}" throughout the rest of the paper. The \texttt{Best} dataset combines the different datasets based on the highest score from different languages. We present the combination of the Best dataset in Table~\ref{tab:data_dec2} and use it to finetune multilingual models as above.

\subsection{Language ID Information}
In a multilingual setting, all sentences of all the languages from Task A are combined to create the training set. To assess the impact of language ID information on the model's performance, we train the model using \textit{tagged} and \textit{untagged} versions based on the five datasets described in Section~\ref{dataset_decision}. Language ID information is not provided in the untagged datasets, which is the same as in the original multilingual dataset. The language ID information is included for the tagged datasets by adding a token denoting the language id at the beginning of each sentence. We evaluate the model on all languages from Task A and their combination. Table~\ref{tab:mul_chs1} presents the performance of our model with tagged and untagged datasets.

We observe that models trained with the tagged dataset generally performed better in the multilingual setting. This indicates that language ID information is helpful for the model to make accurate predictions when dealing with examples from different languages. This observation further supports our decision to use the phylogeny-based adapter, which incorporates language family and genus information into the training.

\begin{table*}[t]\centering
\resizebox{\textwidth}{!}{%
\begin{tabular}{c|ccccc|ccccc}
\toprule
\multirow{2}{*}{\textbf{Lang}} &\textbf{Clean} &\textbf{MT-based} &\textbf{Dict-based} &\textbf{Both} &\textbf{Best} &\textbf{Clean} &\textbf{MT-based} &\textbf{Dict-based} &\textbf{Both} &\textbf{Best} \\
&\textbf{(tagged)} &\textbf{(tagged)} &\textbf{(tagged)} &\textbf{(tagged)} &\textbf{(tagged)} &\textbf{(untagged)} &\textbf{(untagged)} &\textbf{(untagged)} &\textbf{(untagged)} &\textbf{(untagged)} \\\midrule
am &\textbf{64.3} &63.4 &63.5 &63 &63.7 &62.6 &63.1 &62.9 &62.9 &64.2 \\
dz &67 &53.2 &67.1 &49.8 &66.9 &64.3 &50.9 &\textbf{67.4} &50.1 &66.1 \\
ha &\textbf{80.2} &79.3 &79.1 &79.2 &79.8 &79.3 &79.7 &79.2 &78.9 &79.8 \\
ig &75.7 &76.5 &\textbf{77.2} &76.3 &75.8 &73.4 &75.1 &74.7 &74.6 &74.2 \\
kr &71.8 &68 &71.4 &70.5 &71.2 &70.2 &68.9 &\textbf{71.9} &70.3 &70.9 \\
ma &\textbf{78.4} &58.8 &78.2 &59.9 &78.2 &76.7 &59.5 &77.9 &58.3 &77.3 \\
pcm &\textbf{76.8} &57 &76 &55.3 &75.6 &75 &56.4 &74.6 &52.6 &75 \\
pt &\textbf{70.9} &58.7 &56.6 &60.3 &68.6 &70.9 &60.2 &59.5 &60.1 &67.9 \\
sw &63.8 &60.1 &63.5 &60.9 &62.3 &\textbf{64.7} &60.4 &64.4 &59.1 &61.4 \\
ts &50.9 &49.6 &53.1 &52.1 &\textbf{54.4} &52 &50.9 &50.2 &45.4 &49.1 \\
twi &55.8 &23.9 &55.1 &25.5 &57.1 &56.4 &17.8 &\textbf{58.6} & 21.6 &54.5 \\
yo &76.5 &76.3 &76.4 &75.8 &76.6 &76.4 &76.6 &76.5 &\textbf{77.1} &76.7 \\\midrule
multi &\textbf{74.2} &66.7 &73.2 &66.7 &73.8 &72.9 &66.4 &72.7 &65.7 &73 \\
\bottomrule
\end{tabular}
}
\caption{Weighted F$_1$-score on the dev set of our multilingual models on different datasets. 
}
\label{tab:mul_chs1}
\end{table*}

\begin{table*}[h]
\centering
\resizebox{0.8\textwidth}{!}{%
\begin{tabular}{c|ccccccccccccc}
\toprule
\textbf{Model} & \textbf{am} & \textbf{dz} & \textbf{ha} & \textbf{ig} & \textbf{kr} & \textbf{ma} & \textbf{pcm} & \textbf{pt} & \textbf{sw} & \textbf{ts} & \textbf{twi} & \textbf{yo}\\

\midrule
\textbf{Mono} & 63.6 & 58.6 & 79.7 & 74.4 & 67.7 & 77.8 & 76.1 & 64.4 & 62.1 & 51.3 & 51.2 & 74.9 \\
\textbf{Multi} & \textbf{64.3} & \textbf{67.4} & \textbf{80.2} & \textbf{77.2} & \textbf{71.9} & \textbf{78.4} & \textbf{76.8} & \textbf{70.9} & \textbf{64.7} & \textbf{54.4} & \textbf{58.6} & \textbf{77.1} \\

\bottomrule
\end{tabular}
}
\caption{Weighted F$_1$-score on the dev set of monolingual and multilingual model.}
\label{tab:mono_vs_multi}
\end{table*}

In addition, comparing the results in Table~\ref{tab:mono_vs_multi}, we find that the multilingual model outperforms the dedicated monolingual models, suggesting that a multilingual model can more effectively capture sentiment across multiple African languages than one single model. For all future experiments, hence, we only use multilingual models.

\subsection{Phylogeny-based Adaptation}
In all the experiments up until now, we have fine-tuned only the task-adapter. We will now adapter-tune the AfroXLMR-large by inserting phylogeny-based adapter stacks (see Figure~\ref{fig:phylogeny}) inside the Language Model (LM). The intuition behind this is that leveraging the phylogenetic relationships between languages can transfer knowledge and alleviate low-resource scenarios. We will call these adapters family-adapter, genus-adapter, and language-adapter. When adapter-tuning, all the other parameters of the LM will be kept frozen, and only the adapter parameters will be updated through a joint training scheme. Here, we will train the adapters with the Masked Language Modeling task. After the fine-tuning, we will have four adapter components to use in four stack combinations to get our final model. We do not train the task-adapter at this stage but use the previously trained task-adapters.

We benchmark the different combinations as:
\begin{enumerate}[noitemsep,nolistsep]
    \item Task-Adapter
    \item Lang-Adapter + Task Adapter
    \item Genus-Adapter + Lang-Adapter + Task-Adapter
    \item Family-Adapter + Genus-Adapter + Lang-Adapter + Task-Adapter
\end{enumerate}

\begin{table*}[t]\centering
\resizebox{0.75\textwidth}{!}{%
\begin{tabular}{r|lll|lll}
\toprule
\multirow{2}{*}{\textbf{Lang}} &\textbf{Clean} &\textbf{Best} &\textbf{Dict-based} &\textbf{Clean} &\textbf{Both} &\textbf{Dict-based} \\
&\textbf{(tagged)} &\textbf{(tagged)} &\textbf{(tagged)} &\textbf{(untagged)} &\textbf{(untagged)} &\textbf{(untagged)} \\\midrule
am &64.3 \texttt{[T]} &63.8 \texttt{[FGLT]} &64.2 \texttt{[FGLT]} &63.4 \texttt{[FGLT]} &62.9 \texttt{[T]} &62.9 \texttt{[T]} \\
dz &67 \texttt{[T]} &67.1 \texttt{[GLT]} &67.6 \texttt{[GLT]} &64.3 \texttt{[T]} & &67.4 \texttt{[T]} \\
ha &80.2 \texttt{[FGLT]} &79.9 \texttt{[FGLT]} &79.3 \texttt{[FGLT]} &79.8 \texttt{[LT]} &79.7 \texttt{[GLT]} &79.5 \texttt{[LT]} \\
ig &75.7 \texttt{[T]} &76.3 \texttt{[FGLT]} &77.2 \texttt{[T]} &74.1 \texttt{[LT]} &76 \texttt{[LT]} &75.6 \texttt{[LT]} \\
kr &71.8 \texttt{[T]} &72.3 \texttt{[FGLT]} &72.5 \texttt{[FGLT]} &71.9 \texttt{[GLT]} &72.3 \texttt{[LT]} &71.9 \texttt{[T]} \\
ma &78.5 \texttt{[LT]} &78.4 \texttt{[LT]} &78.2 \texttt{[T]} &76.7 \texttt{[T]} & &77.9 \texttt{[T]} \\
pcm &76.8 \texttt{[FGLT]} &75.6 \texttt{[LT]} &76.1 \texttt{[FGLT]} &75.8 \texttt{[LT]} & &75.8 \texttt{[GLT]} \\
pt &70.9 \texttt{[T]} &68.6 \texttt{[T]} & &70.9 \texttt{[T]} & & \\
sw &63.8 \texttt{[T]} &62.5 \texttt{[FGLT]} &63.5 \texttt{[T]} &64.7 \texttt{[T]} &59.3 \texttt{[LT]} &64.4 \texttt{[T]} \\
ts &50.9 \texttt{[T]} &55.1 \texttt{[FGLT]} &53.1 \texttt{[T]} &55.5 \texttt{[FGLT]} &49.3 \texttt{[FGLT]} &50.2 \texttt{[T]} \\
twi &56.3 \texttt{[GLT]} &57.7 \texttt{[LT]} &55.1 \texttt{[T]} &56.4 \texttt{[T]} & &58.6 \texttt{[T]} \\
yo &76.5 \texttt{[T]} &76.9 \texttt{[LT]} &76.4 \texttt{[T]} &76.4 \texttt{[T]} &77.1 \texttt{[T]} &76.5 \texttt{[T]} \\
\bottomrule
\end{tabular}
}
\caption{Weighted F$_1$-score on the dev set of our Phylogeny-based models trained with the Clean dataset. The Column indicates which task-adapter has been used. [n] indicates which configuration gives us the best F$_1$-score. \texttt{[T]} = Task, \texttt{[LT]} = Lang + Task, \texttt{[GLT]} = Genus + Lang + Task, \texttt{[FGLT]} = Family + Genus + Lang + Task}
\label{tab:phy_ada1}
\end{table*}

Table~\ref{tab:phy_ada1} shows the results with the best out of the above four configurations. We adapter-tune the MLM using the Clean dataset. For task-adapter we use the ones that got the highest scores in Table \ref{tab:mul_chs1} for different languages. We also adapter-tune the MLM using the Clean + Dictionary-based, Clean + Both, and Best datasets. See Appendix \ref{sec:adapter-tuning} for all the results.

\section{Results and Discussion}
\paragraph{Monolingual Performance}
For Task A, we ensemble the best five models we found for each language. We use majority voting to obtain the final output, resolving ties randomly.

Our system's test performance in 12 languages is presented in Table~\ref{tab:main_result}. Notably, our system demonstrates impressive results on Track 5: Amharic, ranking first on the leaderboard and outperforming the second-place system by a significant margin of 6.2 points. In addition, our system falls only 0.8 points short of the top-performing system in Track 9: Kinyarwanda. In both Track 5 and Track 9, the best data are from Clean and Dictionary-based. While our system does not achieve the top performance for other tracks, it still achieves highly competitive results across all tracks, with a top-10 ranking in 4 out of the 12 tracks.

Our success in fine-tuning AfroXLMR-large can be attributed to several factors. First, AfroXLMR-large is a powerful pre-trained language model specifically designed for African languages. This provides an excellent starting point for fine-tuning specific tasks. Second, we carefully selected the best dataset for each language, which ensures that our system is trained on the most appropriate data for the particular language.

\paragraph{Multilingual Performance}
For Task B, we can not follow the same procedure as in Task A, as the language ID information is absent; we cannot utilize phylogeny-based adapter-tuning to enable information sharing between similar languages during inference time. We could have used a Lang ID model but chose not to because langID for African languages is untrustworthy. So, we use the best model from Table~\ref{tab:mul_chs1} for this task. Table \ref{tab:main_result} presents the performance of our system, which falls only 3.82 points short of the top-performing system, achieving a respectable 7th place ranking. While it is not ideal, it still demonstrates the potential of our approach in real-world scenarios where language ID information is often missing.

One possible reason for the suboptimal performance could be the limitations of the available data. Although we carefully select the best dataset from each language, the amount and quality of the data are still limited. In addition, the absence of language ID information makes it more challenging to distinguish between closely related languages, which can lead to a higher degree of semantic ambiguity and reduce the accuracy of our system.

\paragraph{Zero-Shot Performance}
For Task C, we use the same model as Task B. Table~\ref{tab:main_result} also shows the zero-shot performance of our system on Track 17: Tigrinya and Track 18: Oromo. Our system ranks 11th and 13th on the leaderboards, falling short of the top-performing models by 9.34 and 4.36 points, respectively. It highlights the potential of our system for zero-shot cross-lingual transfer learning, although further improvements may be needed to achieve state-of-the-art performance in these languages. For example, we could have used additional unlabeled data in the two languages to continue training the base model as \cite{muller-etal-2021-unseen}. We leave these explorations for future work.

\begin{table}[t]\centering
\resizebox{0.45\textwidth}{!}{%
\begin{tabular}{c|c|c|c}
\toprule
\textbf{Task} & \textbf{Lang}  & \textbf{Weighted F$_1$($\Delta$)} & \textbf{Ranking}  \\
\midrule
\multirow{12}{*}{Task A} & ha & 79.6 (-3.1)  & 17/35\\
& yo & 70.8 (-9.3)  & 21/33\\
& ig & 75.3 (-7.6)  & 24/32\\
& pcm & 68.8 (-7.1)  & 11/32\\
& \textbf{am} & \textbf{78.4 (0)}  & 1/29\\
& dz & 68 (-6.2)  & 15/30\\
& ma & 55.2 (-9.6)  & 19/32\\
& sw & 63.7 (-2)  & 6/30\\ 
& \textbf{kr} & \textbf{71.8 (-0.8)}  & 5/34\\
& twi & 56.5 (-11.8)  & 28/31\\ 
& pt & 71.9 (-3.1)  & 10/30\\ 
& ts & 51.7 (-9)  & 15/31\\ 
\midrule
Task B & multi & 71.2 (-3.8)  & 7/33\\ 
\midrule
\multirow{2}{*}{Task C} & tg & 61.5 (-9.3)  & 11/29\\
& or & 41.9 (-4.4)  & 13/27\\
\bottomrule
\end{tabular}
}
\caption{Weighted F$_1$-score on the test set of the 15 tracks we participated in. $\Delta$ shows the offset of our scores from the best-performing system for each track. Highlighted are the ones that have $\Delta$ smaller than -1.}
\label{tab:main_result}
\end{table}

\section{Conclusion}
This paper describes GMU’s SA systems for the AfriSenti SemEval-2023 shared task. We participated in all three sub-tasks: Monolingual, Multilingual, and Zero-Shot. As a starting point for our system, we leverage AfroXLMR-large, a pre-trained multilingual language model specifically trained on African languages and then fine-tuned with original and augmented training data. To further enhance our system, we utilize phylogeny-based adapter-tuning, which involves adapting to the target languages using knowledge from related languages in the phylogenetic tree. Multiple models are created and ensembled to obtain the best results. Our system outperforms all other systems in track 5: Amharic, achieving the highest F$_1$-score with a remarkable 6.2-point higher than the second-best performing system.

\section*{Acknowledgements}
 This work was generously supported by NSF awards IIS-2125466 and 2040926 and a Sponsored Research Award from Meta. We thank the GMU Office of Research Computing for granting us access to large computing resources.

\bibliography{anthology,custom}

\begin{thebibliography}{28}
\expandafter\ifx\csname natexlab\endcsname\relax\def\natexlab#1{#1}\fi

\bibitem[{Alabi et~al.(2022)Alabi, Adelani, Mosbach, and
  Klakow}]{alabi-etal-2022-adapting}
Jesujoba~O. Alabi, David~Ifeoluwa Adelani, Marius Mosbach, and Dietrich Klakow.
  2022.
\newblock \href {https://aclanthology.org/2022.coling-1.382} {Adapting
  pre-trained language models to {A}frican languages via multilingual adaptive
  fine-tuning}.
\newblock In \emph{Proceedings of the 29th International Conference on
  Computational Linguistics}, pages 4336--4349, Gyeongju, Republic of Korea.
  International Committee on Computational Linguistics.

\bibitem[{Alam and Anastasopoulos(2022)}]{alam-anastasopoulos:2022:WMT}
Md~Mahfuz~Ibn Alam and Antonios Anastasopoulos. 2022.
\newblock \href {https://aclanthology.org/2022.wmt-1.99} {Language adapters for
  large-scale mt: The gmu system for the wmt 2022 large-scale machine
  translation evaluation for african languages shared task}.
\newblock In \emph{Proceedings of the Seventh Conference on Machine
  Translation}, pages 1015--1033, Abu Dhabi. Association for Computational
  Linguistics.

\bibitem[{Ansell et~al.(2021)Ansell, Ponti, Pfeiffer, Ruder, Glava{\v{s}},
  Vuli{\'c}, and Korhonen}]{ansell-etal-2021-mad-g}
Alan Ansell, Edoardo~Maria Ponti, Jonas Pfeiffer, Sebastian Ruder, Goran
  Glava{\v{s}}, Ivan Vuli{\'c}, and Anna Korhonen. 2021.
\newblock \href {https://doi.org/10.18653/v1/2021.findings-emnlp.410}
  {{MAD}-{G}: {M}ultilingual adapter generation for efficient cross-lingual
  transfer}.
\newblock In \emph{Findings of the Association for Computational Linguistics:
  EMNLP 2021}, pages 4762--4781, Punta Cana, Dominican Republic. Association
  for Computational Linguistics.

\bibitem[{Blasi et~al.(2022)Blasi, Anastasopoulos, and
  Neubig}]{blasi-etal-2022-systematic}
Damian Blasi, Antonios Anastasopoulos, and Graham Neubig. 2022.
\newblock \href {https://doi.org/10.18653/v1/2022.acl-long.376} {Systematic
  inequalities in language technology performance across the world{'}s
  languages}.
\newblock In \emph{Proceedings of the 60th Annual Meeting of the Association
  for Computational Linguistics (Volume 1: Long Papers)}, pages 5486--5505,
  Dublin, Ireland. Association for Computational Linguistics.

\bibitem[{Chronopoulou et~al.(2022)Chronopoulou, Peters, and
  Dodge}]{chronopoulou-etal-2022-efficient}
Alexandra Chronopoulou, Matthew Peters, and Jesse Dodge. 2022.
\newblock \href {https://doi.org/10.18653/v1/2022.naacl-main.96} {Efficient
  hierarchical domain adaptation for pretrained language models}.
\newblock In \emph{Proceedings of the 2022 Conference of the North American
  Chapter of the Association for Computational Linguistics: Human Language
  Technologies}, pages 1336--1351, Seattle, United States. Association for
  Computational Linguistics.

\bibitem[{Conneau et~al.(2019)Conneau, Khandelwal, Goyal, Chaudhary, Wenzek,
  Guzm{\'{a}}n, Grave, Ott, Zettlemoyer, and
  Stoyanov}]{DBLP:journals/corr/abs-1911-02116}
Alexis Conneau, Kartikay Khandelwal, Naman Goyal, Vishrav Chaudhary, Guillaume
  Wenzek, Francisco Guzm{\'{a}}n, Edouard Grave, Myle Ott, Luke Zettlemoyer,
  and Veselin Stoyanov. 2019.
\newblock \href {http://arxiv.org/abs/1911.02116} {Unsupervised cross-lingual
  representation learning at scale}.
\newblock \emph{CoRR}, abs/1911.02116.

\bibitem[{Devlin et~al.(2018)Devlin, Chang, Lee, and
  Toutanova}]{DBLP:journals/corr/abs-1810-04805}
Jacob Devlin, Ming{-}Wei Chang, Kenton Lee, and Kristina Toutanova. 2018.
\newblock \href {http://arxiv.org/abs/1810.04805} {{BERT:} pre-training of deep
  bidirectional transformers for language understanding}.
\newblock \emph{CoRR}, abs/1810.04805.

\bibitem[{Dossou et~al.(2022)Dossou, Tonja, Yousuf, Osei, Oppong, Shode,
  Awoyomi, and Emezue}]{dossou-etal-2022-afrolm}
Bonaventure F.~P. Dossou, Atnafu Tonja, Oreen Yousuf, Salomey Osei, Abigail
  Oppong, Iyanuoluwa Shode, Oluwabusayo~Olufunke Awoyomi, and Chris Emezue.
  2022.
\newblock \href {https://aclanthology.org/2022.sustainlp-1.11} {{A}fro{LM}: A
  self-active learning-based multilingual pretrained language model for 23
  {A}frican languages}.
\newblock In \emph{Proceedings of The Third Workshop on Simple and Efficient
  Natural Language Processing (SustaiNLP)}, pages 52--64, Abu Dhabi, United
  Arab Emirates (Hybrid). Association for Computational Linguistics.

\bibitem[{Faisal and
  Anastasopoulos(2022)}]{faisal-anastasopoulos-2022-phylogeny}
Fahim Faisal and Antonios Anastasopoulos. 2022.
\newblock \href {https://aclanthology.org/2022.aacl-main.34}
  {Phylogeny-inspired adaptation of multilingual models to new languages}.
\newblock In \emph{Proceedings of the 2nd Conference of the Asia-Pacific
  Chapter of the Association for Computational Linguistics and the 12th
  International Joint Conference on Natural Language Processing (Volume 1: Long
  Papers)}, pages 434--452, Online only. Association for Computational
  Linguistics.

\bibitem[{Houlsby et~al.(2019)Houlsby, Giurgiu, Jastrzebski, Morrone,
  De~Laroussilhe, Gesmundo, Attariyan, and Gelly}]{houlsby2019parameter}
Neil Houlsby, Andrei Giurgiu, Stanislaw Jastrzebski, Bruna Morrone, Quentin
  De~Laroussilhe, Andrea Gesmundo, Mona Attariyan, and Sylvain Gelly. 2019.
\newblock Parameter-efficient transfer learning for nlp.
\newblock In \emph{International Conference on Machine Learning}, pages
  2790--2799. PMLR.

\bibitem[{Joshi et~al.(2020)Joshi, Santy, Budhiraja, Bali, and
  Choudhury}]{joshi-etal-2020-state}
Pratik Joshi, Sebastin Santy, Amar Budhiraja, Kalika Bali, and Monojit
  Choudhury. 2020.
\newblock \href {https://doi.org/10.18653/v1/2020.acl-main.560} {The state and
  fate of linguistic diversity and inclusion in the {NLP} world}.
\newblock In \emph{Proceedings of the 58th Annual Meeting of the Association
  for Computational Linguistics}, pages 6282--6293, Online. Association for
  Computational Linguistics.

\bibitem[{Kamholz et~al.(2014)Kamholz, Pool, and
  Colowick}]{kamholz-etal-2014-panlex}
David Kamholz, Jonathan Pool, and Susan Colowick. 2014.
\newblock \href
  {http://www.lrec-conf.org/proceedings/lrec2014/pdf/1029_Paper.pdf}
  {{P}an{L}ex: Building a resource for panlingual lexical translation}.
\newblock In \emph{Proceedings of the Ninth International Conference on
  Language Resources and Evaluation ({LREC}'14)}, pages 3145--3150, Reykjavik,
  Iceland. European Language Resources Association (ELRA).

\bibitem[{Kingma and Ba(2014)}]{kingma2014adam}
Diederik~P Kingma and Jimmy Ba. 2014.
\newblock Adam: A method for stochastic optimization.
\newblock ArXiv:1412.6980.

\bibitem[{Liu et~al.(2019)Liu, Ott, Goyal, Du, Joshi, Chen, Levy, Lewis,
  Zettlemoyer, and Stoyanov}]{liu2019roberta}
Yinhan Liu, Myle Ott, Naman Goyal, Jingfei Du, Mandar Joshi, Danqi Chen, Omer
  Levy, Mike Lewis, Luke Zettlemoyer, and Veselin Stoyanov. 2019.
\newblock Roberta: A robustly optimized bert pretraining approach.
\newblock ArXiv:1907.11692.

\bibitem[{Mabokela and Schlippe(2022)}]{mabokela2022sentiment}
Ronny Mabokela and Tim Schlippe. 2022.
\newblock A sentiment corpus for south african under-resourced languages in a
  multilingual context.
\newblock In \emph{Proceedings of the 1st Annual Meeting of the ELRA/ISCA
  Special Interest Group on Under-Resourced Languages}, pages 70--77.

\bibitem[{Martin et~al.(2022)Martin, Mswahili, Jeong, and
  Young-Seob}]{martin2022swahbert}
Gati Martin, Medard~Edmund Mswahili, Young-Seob Jeong, and Jeong Young-Seob.
  2022.
\newblock Swahbert: Language model of swahili.
\newblock In \emph{Proceedings of the 2022 Conference of the North American
  Chapter of the Association for Computational Linguistics: Human Language
  Technologies}, pages 303--313.

\bibitem[{Martin et~al.(2021)Martin, Mswahili, and Jeong}]{martin2021sentiment}
Gati~L Martin, Medard~E Mswahili, and Young-Seob Jeong. 2021.
\newblock Sentiment classification in swahili language using multilingual bert.
\newblock ArXiv:2104.09006.

\bibitem[{Mohammad(2022)}]{mohammad2022ethics}
Saif~M Mohammad. 2022.
\newblock Ethics sheet for automatic emotion recognition and sentiment
  analysis.
\newblock \emph{Computational Linguistics}, 48(2):239--278.

\bibitem[{Muhammad et~al.(2023{\natexlab{a}})Muhammad, Abdulmumin, Ayele,
  Ousidhoum, Adelani, Yimam, Ahmad, Beloucif, Mohammad, Ruder, Hourrane,
  Brazdil, Ali, David, Osei, Bello, Ibrahim, Gwadabe, Rutunda, Belay, Messelle,
  Balcha, Chala, Gebremichael, Opoku, and Arthur}]{muhammad2023afrisenti}
Shamsuddeen~Hassan Muhammad, Idris Abdulmumin, Abinew~Ali Ayele, Nedjma
  Ousidhoum, David~Ifeoluwa Adelani, Seid~Muhie Yimam, Ibrahim~Sa'id Ahmad,
  Meriem Beloucif, Saif~M. Mohammad, Sebastian Ruder, Oumaima Hourrane, Pavel
  Brazdil, Felermino Dário Mário~António Ali, Davis David, Salomey Osei,
  Bello~Shehu Bello, Falalu Ibrahim, Tajuddeen Gwadabe, Samuel Rutunda, Tadesse
  Belay, Wendimu~Baye Messelle, Hailu~Beshada Balcha, Sisay~Adugna Chala,
  Hagos~Tesfahun Gebremichael, Bernard Opoku, and Steven Arthur.
  2023{\natexlab{a}}.
\newblock \href {https://doi.org/10.48550/arXiv.2302.08956} {{AfriSenti: A
  Twitter Sentiment Analysis Benchmark for African Languages}}.

\bibitem[{Muhammad et~al.(2023{\natexlab{b}})Muhammad, Abdulmumin, Yimam,
  Adelani, Ahmad, Ousidhoum, Ayele, Mohammad, Beloucif, and
  Ruder}]{muhammadSemEval2023}
Shamsuddeen~Hassan Muhammad, Idris Abdulmumin, Seid~Muhie Yimam, David~Ifeoluwa
  Adelani, Ibrahim~Sa'id Ahmad, Nedjma Ousidhoum, Abinew~Ali Ayele, Saif~M.
  Mohammad, Meriem Beloucif, and Sebastian Ruder. 2023{\natexlab{b}}.
\newblock {SemEval-2023 Task 12: Sentiment Analysis for African Languages
  (AfriSenti-SemEval)}.
\newblock In \emph{Proceedings of the 17th {{International Workshop}} on
  {{Semantic Evaluation}} ({{SemEval-2023}})}. {Association for Computational
  Linguistics}.

\bibitem[{Muhammad et~al.(2022)Muhammad, Adelani, Ruder, Ahmad, Abdulmumin,
  Bello, Choudhury, Emezue, Abdullahi, Aremu, Jorge, and
  Brazdil}]{muhammad-etal-2022-naijasenti}
Shamsuddeen~Hassan Muhammad, David~Ifeoluwa Adelani, Sebastian Ruder,
  Ibrahim~Sa{'}id Ahmad, Idris Abdulmumin, Bello~Shehu Bello, Monojit
  Choudhury, Chris~Chinenye Emezue, Saheed~Salahudeen Abdullahi, Anuoluwapo
  Aremu, Al{\'\i}pio Jorge, and Pavel Brazdil. 2022.
\newblock \href {https://aclanthology.org/2022.lrec-1.63} {{N}aija{S}enti: A
  {N}igerian {T}witter sentiment corpus for multilingual sentiment analysis}.
\newblock In \emph{Proceedings of the Thirteenth Language Resources and
  Evaluation Conference}, pages 590--602, Marseille, France. European Language
  Resources Association.

\bibitem[{Muller et~al.(2021)Muller, Anastasopoulos, Sagot, and
  Seddah}]{muller-etal-2021-unseen}
Benjamin Muller, Antonios Anastasopoulos, Beno{\^\i}t Sagot, and Djam{\'e}
  Seddah. 2021.
\newblock \href {https://doi.org/10.18653/v1/2021.naacl-main.38} {When being
  unseen from m{BERT} is just the beginning: Handling new languages with
  multilingual language models}.
\newblock In \emph{Proceedings of the 2021 Conference of the North American
  Chapter of the Association for Computational Linguistics: Human Language
  Technologies}, pages 448--462, Online. Association for Computational
  Linguistics.

\bibitem[{Ogueji et~al.(2021)Ogueji, Zhu, and Lin}]{ogueji-etal-2021-small}
Kelechi Ogueji, Yuxin Zhu, and Jimmy Lin. 2021.
\newblock \href {https://aclanthology.org/2021.mrl-1.11} {Small data? no
  problem! exploring the viability of pretrained multilingual language models
  for low-resourced languages}.
\newblock In \emph{Proceedings of the 1st Workshop on Multilingual
  Representation Learning}, pages 116--126, Punta Cana, Dominican Republic.
  Association for Computational Linguistics.

\bibitem[{Shode et~al.(2022)Shode, Adelani, and Feldman}]{shode2022yosm}
Iyanuoluwa Shode, David~Ifeoluwa Adelani, and Anna Feldman. 2022.
\newblock \textsc{yosm}: A new yoruba sentiment corpus for movie reviews.
\newblock ArXiv:2204.09711.

\bibitem[{Skutnabb-Kangas et~al.(2003)Skutnabb-Kangas, Maffi, and
  Harmon}]{skutnabb2003sharing}
Tove Skutnabb-Kangas, Luisa Maffi, and David Harmon. 2003.
\newblock \emph{Sharing a world of difference: the earth's linguistic, cultural
  and biological diversity}.
\newblock Unesco.

\bibitem[{Socher et~al.(2013)Socher, Perelygin, Wu, Chuang, Manning, Ng, and
  Potts}]{socher-etal-2013-recursive}
Richard Socher, Alex Perelygin, Jean Wu, Jason Chuang, Christopher~D. Manning,
  Andrew Ng, and Christopher Potts. 2013.
\newblock \href {https://aclanthology.org/D13-1170} {Recursive deep models for
  semantic compositionality over a sentiment treebank}.
\newblock In \emph{Proceedings of the 2013 Conference on Empirical Methods in
  Natural Language Processing}, pages 1631--1642, Seattle, Washington, USA.
  Association for Computational Linguistics.

\bibitem[{Xia et~al.(2019)Xia, Kong, Anastasopoulos, and
  Neubig}]{xia-etal-2019-generalized}
Mengzhou Xia, Xiang Kong, Antonios Anastasopoulos, and Graham Neubig. 2019.
\newblock \href {https://doi.org/10.18653/v1/P19-1579} {Generalized data
  augmentation for low-resource translation}.
\newblock In \emph{Proceedings of the 57th Annual Meeting of the Association
  for Computational Linguistics}, pages 5786--5796, Florence, Italy.
  Association for Computational Linguistics.

\bibitem[{Xie and Anastasopoulos(2023)}]{xie-anastasopoulos-23-noisy}
Ruoyu Xie and Antonios Anastasopoulos. 2023.
\newblock \href {https://arxiv.org/abs/2301.09685} {Noisy parallel data
  alignment}.
\newblock In \emph{Findings of the 17th Conference of the European Chapter of
  the Association for Computational Linguistics (EACL 2023)}, Dubrovnik,
  Croatia. Association for Computational Linguistics.

\end{thebibliography}
\bibliographystyle{acl_natbib}

\clearpage
\newpage
\appendix
\onecolumn
\section{Best Dataset}
\begin{table*}[h!]\centering
\resizebox{0.35\textwidth}{!}{%
\begin{tabular}{c|c}
\toprule
\textbf{Lang}  & \textbf{Dataset} \\
\midrule
ha & Clean\\
yo & Clean + MT-based\\
ig & Clean + Both\\
pcm & Clean + Dictionary-based\\
am & Clean + Dictionary-based\\
dz & Clean + Dictionary-based\\
ma & Clean + Dictionary-based\\
sw & Clean + MT-based\\ 
kr & Clean + Dictionary-based\\
twi & Clean + Dictionary-based\\ 
pt & Clean\\ 
ts & Clean + MT-based\\
\bottomrule
\end{tabular}
}
\caption{The dataset combination of the Best dataset.}
\label{tab:data_dec2}
\end{table*}

\section{Adapter-tuning}\label{sec:adapter-tuning}
Tables~\ref{tab:ada_tune1} and~\ref{tab:ada_tune2} present adapter-tuning results on Dict-based datasets, Best, and Both datasets, respectively.

\begin{table*}[h!]\centering
\resizebox{0.75\textwidth}{!}{%
\begin{tabular}{c|ccc|ccc}
\toprule
\multirow{2}{*}{\textbf{Lang}} &\textbf{Clean} &\textbf{Best} &\textbf{Dict-based} &\textbf{Clean} &\textbf{Both} &\textbf{Dict-based} \\
&\textbf{(tagged)} &\textbf{(tagged)} &\textbf{(tagged)} &\textbf{(untagged)} &\textbf{(untagged)} &\textbf{(untagged)} \\\midrule	
am &64.3 [1] &63.6 [1] &64 [4] &62.6 [1] &63.3 [4] &62.9 [1] \\
dz &67.7 [2] &68.2 [3] &68.5 [4] &64.4 [3] &- &67.4 [1] \\
ha &80.2 [1] &79.8 [1] &79.3 [2] &79.5 [2] &79.8 [2] &79.5 [2] \\
ig &76.1 [2] &75.8 [1] &77.2 [1] &73.8 [2] &74.8 [2] &74.7 [1] \\
kr &71.8 [1] &71.2 [2] &71.6 [3] &71.0 [4] &70.9 [2] &71.9 [1] \\
ma &78.4 [1] &79.3 [2] &78.2 [1] &76.7 [1] &- &77.9 [1] \\
pcm &76.8 [4] &75.6 [1] &76.1 [4] &76.0 [4] &- &75.9 [3] \\
pt &-   & - &-   &  - &-   & - \\
sw &64 [1] &62.8 [43] &63.5 [1] &64.7 [1] &59.1 [1] &64.4 [1] \\
ts &50.9 [1] &54.4 [1] &53.1 [1] &57.1 [2] &50.1 [2] &50.2 [1] \\
twi &55.8 [2] &57.7 [2] &55.1 [1] &56.8 [2] & -&58.6 [1] \\
yo &76.5 [1] &76.9 [2] &76.4 [1] &76.4 [1] &77.1 [1] &76.5 [1] \\
\bottomrule
\end{tabular}
}
\caption{Weighted F$_1$-score on the dev set of our Phylogeny-based models trained with \textbf{Dict-based} datasets. The column indicates which task-adapter has been used. [n] indicates which configuration gives us the best F$_1$-score. [1] = Task, [2] = Lang + Task, [3] = Genus + Lang + Task, [4] = Family + Genus + Lang + Task}
\label{tab:ada_tune1}
\end{table*}

\begin{table}[h!]
\centering
\begin{tabular}{cc}
\resizebox{0.45\textwidth}{!}{%
\begin{tabular}{c|ccc}
\toprule
\multirow{2}{*}{\textbf{Lang}} &\textbf{Clean} &\textbf{Best} &\textbf{Dict-based} \\
&\textbf{(tagged)} &\textbf{(tagged)} &\textbf{(tagged)}  \\\midrule	
am &64.3 [1] &63.6 [1] &64.4 [3] \\
dz &67.1 [2] &67.4 [3] &68.2 [4] \\
ha &80.2 [3] &80 [3] &79.3 [4] \\
ig &75.7 [1] &75.8 [1] &77.2 [1]  \\
kr &71.8 [1] &71.2 [1] &71.4 [1] \\
ma &78.4 [1] &78.2 [1] &78.2 [1] \\
pcm &76.8 [4] &75.6 [1] &76.1 [4]  \\
pt &70.9 [1]  & 68.6 [1] & -  \\
sw &63.8 [1] &62.4 [2] &63.5 [1] \\
ts &50.9 [1] &57.3 [3] &53.1 [1]  \\
twi &55.8 [2] &57.5 [2] &55.1 [1] \\
yo &76.5 [1] &76.2 [2] &76.4 [1]  \\
\bottomrule
\end{tabular}
}
%\label{tab:ada_tune2}
&
\resizebox{0.45\textwidth}{!}{%
\begin{tabular}{c|ccc}
\toprule
\multirow{2}{*}{\textbf{Lang}} &\textbf{Clean} &\textbf{Best} &\textbf{Dict-based} \\
&\textbf{(tagged)} &\textbf{(tagged)} &\textbf{(tagged)}  \\\midrule	
am &64.1 [3] &62.9 [1] &63.2 [2] \\
dz & - & - & -\\
ha &80.2 [2] &79.8 [4] &79.7 [2] \\
ig &74.1 [2] &75.4 [2] &74.7 [1]  \\
kr &71.6 [3] &71.3 [2] &72.3 [2] \\
ma  & - & - & -\\
pcm  & - & - & -\\
pt &-   & - &-  \\
sw &64.8 [3] &59.1 [1] &64.1 [1] \\
ts &56.5 [4] &52.4 [3] &53.6 [2]  \\
twi & - & - & -\\
yo &76.4 [1] &77.1 [1] &76.6 [2]  \\
\bottomrule
\end{tabular}
}
%\label{tab:ada_tune3} 
\\
Results using the \texttt{Best} dataset.
&
Results using the \texttt{Both} dataset.
%Weighted F$_1$-score on the dev set of our Phylogeny-based models trained with \textbf{Both} datasets. The column indicates which task-adapter has been used. [n] indicates which configuration gives us the best F$_1$-score. [1] = Task, [2] = Lang + Task, [3] = Genus + Lang + Task, [4] = Family + Genus + Lang + Task.
\end{tabular}
\caption{Weighted F$_1$-score on the dev set of our Phylogeny-based models trained with \texttt{Best} (left) or \texttt{Both} (right) datasets. The column indicates which task-adapter has been used. [n] indicates which configuration gives us the best F$_1$-score. [1] = Task, [2] = Lang + Task, [3] = Genus + Lang + Task, [4] = Family + Genus + Lang + Task.}
\label{tab:ada_tune2}
\end{table}

\end{document}